\documentclass{article}
\usepackage[utf8]{inputenc}
\usepackage{graphicx}
\newtheorem{prop}{Proposition}
\newtheorem{cor}{Corollary}
\newtheorem{rem}{Remark}
\newtheorem{adv}{Advantage}
\newtheorem{contr}{Contribution}
\title{Statistical Formulas for F Measures}
\author{Wenxin Jiang\\Northwestern University, USA \\ wjiang@northwestern.edu}
\date{December 2020}

\begin{document}

\maketitle

Abstract: We provide analytic formulas for the standard error and confidence intervals for the F measures, based on a property of asymptotic normality in the large sample limit. The formula can be applied for sample size planning in order to achieve accurate enough estimation of these F measures.


MSC-class:	62F12, 62P99

\section{Introduction}
The F measures are very commonly used to estimate the performance of machine learning methods (see, e.g., the Wikipedia entry of F score). This paper provides simple formulas for their standard errors, probability distributions, and the related confidence intervals and sample size planning based on large data.
We will first use a real data set (Stine,  Foster,  and Waterman 1998) to illustrate the concept of the F  measures.
A purchase for one of the two brands of orange juices: Citrus Hill and Minimaid, is coded respectively as $Z=1$ and $Z=0$ and modeled as a random variable. A score  $S$ summarizing the preference to the Citrus Hill brand is assigned to this purchase. This score $S$ is also modeled as a random variable since it depends on factors such as customer loyalty and price difference, which can differ for each purchase. 
(See, e.g., Jiang and Zhao 2015, who obtain $S$ by logistic regression  from a training data set.)  Then $n=535$  purchases in a  validation data set can be used to evaluate a prediction rule of the form $A=I(S>t)$ for some threshold $t$, for the purpose of predicting $Z$.\footnote{$I(\, )$ is an indicator function defined as $I(event)=1$ if the event happens, or else $I(event)=0$.} 
 Three common metrics of interest are precision
$prec =P(Z=1|A=1)$ (which is also $E(ZA)/EA$), recall $rec=P(A=1|Z=1)$ (which is also $E(ZA)/EZ$),\footnote{The symbols $E$ and $P$ represent expectation and probability, respectively.  So the precision represents the proportion of predicted Citrus Hill purchases that  are  true Citrus Hill purchases. The recall represents the proportion of true Citrus Hill purchases that are predicted to be Citrus Hill purchases.} and the F-beta  measure
$$
\phi_\beta=(1+\beta^2)/( 1/prec+\beta^2/rec),  
$$
for some parameter $\beta$, that adjusts the relative importance of recall and precision. Commonly used values include $\beta \in\{0.5,1,2\}$.

All these measures are valued in $[0,1]$, and the higher the better.
Note that one can easily increase  recall to 1 by decreasing the threshold $t$ in the  prediction rule $A=I(S>t)$, but this usually will decrease the precision. On the other hand, improving the precision might hurt the recall. The  F measure,  compromising both precision and recall,  then naturally becomes very popular.

 In the orange juice data set above, the score $S$ models the probability of $Z=1$ based on a logistic regression, so we will use  $A=I(S>0.5)$ as the prediction rule for $Z$.
 The F measure is estimated by a sample version, where expectation $E$ is replaced by a sample average $E_n$ over $n$ purchases in the validation data set. The estimate, e.g., for the F0.5 measure $\phi_{0.5}$, is 0.861. However, this value is only based on a sample $n=535$. 
 
 {\em What is the standard error? What is a confidence interval?} These are natural questions that we want to answer.
 
Previous works on this include Bayesian methods (e.g., Goutte and  Gaussier 2005), cross validation (e.g., Wang, Li, Li, Wang and Yang, 2015) and   bootstrap methods (e.g., Itzikovitch 2019), but no one uses analytic formulas for the standard error. However, for the special case of F1 measure (where $\beta=1$), such an analytic formula  in fact exists (Janson  and Vegelius 1981, Elston,   Schroeder, and Rohjan  1982), but only is largely unknown due to a combination of two facts:   the fact that F1 was named differently (as the Dice coefficient, Dice 1945), and that the analytic formula for its standard error first appeared in a very different field (ecology).

Our paper makes two contributions:

\begin{contr}
 We bring the analytic approach of   Janson  and Vegelius (1981)  and Elston et al. (1982) to the attention of the machine learning community, and generalize it for the standard error of the F-beta measure with any  importance parameter $\beta$.  \end{contr}

\begin{contr} We study how to use these analytic formulas to plan sample size in order to achieve accurate enough estimation of the F measures.\end{contr}

 Our results actually hold for a more general measure called the Tversky index that appeared in the field of  psychology  (Tversky 1977). 
 
\section{Main results}
 \subsection{Formulas}
 
 \begin{prop}
 Assume that $(Z,A),(Z_1,A_1),(Z_2,A_2),...,(Z_n,A_n),...$ are iid (independent and identically distributed) random variables on $\{0,1\}\times\{0,1\}$. Denote the sample average by
 $ E_n f(Z,A)\equiv n^{-1}\sum_{i=1}^n f(Z_i,A_i)$ and symbolically let $E_\infty=E$.
 For any $a,b>0$, $n=1,2,...,\infty$, let 
 $$\tau_n(a,b)\equiv \frac{E_n(ZA)}{E_n(ZA)+aE_n(A(1-Z))+b E_n(Z(1-A))}
$$ be the Tversky index, 
and denote $$
\nu_n(a,b) \equiv  \frac{[\tau_n(a^2,b^2)]^{-1}-1+\{[\tau_n(a,b)]^{-1}-1\}^2}{E_n(ZA)[\tau_n(a,b)]^{-4}} ,$$

and symbolically let $\tau(a,b)_\infty =\tau(a,b)$ and $\nu(a,b)_\infty=\nu(a,b)$.

 Then we have:
 
 (i)
 $\sqrt{n}(\tau_n(a,b)-\tau(a,b))$ converges in distribution to $N(0, \nu(a,b))$ as $n\rightarrow\infty$.
 
 (ii) The variance has a common upper bound for any prediction rule $A$: 
 $$
 \nu(a,b)\leq  \frac{V(\max\{a,b\})}{ bEZ },
 $$
 where
 
 $V(\max\{a,b\}))= \tau_o(1-\tau_o)(1- \tau_o/c)^2$, 
 
  $\tau_o=\tau_-I(c>1) +\tau_+ I(c<0)$, $\tau_\pm=(3+2c\pm \sqrt{4c^2-4c+9})/8$,

 $c=1/(1-\max\{a,b\})$.
 
(iii) Some potentially useful values of  $V(\cdot)$ are provided here:
 \begin{table}[h]
\begin{tabular}{llllll}
 $\max\{a,b\}$&.5&.6  &.7  &.8  &.9  \\
 $V(\max\{a,b\}))$& .1549& .1695 &.1861&  .2050& .2262 \\
%
\end{tabular}
\end{table}
 \end{prop}

\begin{rem}
The above proposition is for the Tversky index, of which the F measure is a special case, where 
$\phi_\beta=\tau(a,b)$,
$a= (1+\beta^2)^{-1}$, $b=\beta^2(1+\beta^2)^{-1}$.\end{rem}

So $\phi_{0.5}=\tau(0.8, 0.2)$, $\phi_{1}=\tau(0.5, 0.5)$, $\phi_{2}=\tau(0.2, 0.8)$.

\begin{rem}The formula of $\nu_n(a,b)$ in general depends on three quantities $E_nZA$, $\tau_n(a,b)$ and $\tau_n(a^2,b^2)$. However, when $a=b$ such as for F1 measure $\tau_n(0.5,0.5)$, the dependence of $\nu_n(a,b)$ on  $\tau_n(a^2,b^2)$ can be removed by using a relation 
$[\tau_n(a^2,a^2)]^{-1}-1= a\{[\tau_n(a,a)]^{-1}-1\}$.\end{rem}
\begin{cor}

The proposition implies that, for any $\alpha \in(0,1)$,
 $$
 \lim_{n\rightarrow \infty}P[\tau(a,b)\in  \tau_n(a,b)\pm \Phi^{-1}(1-\frac{\alpha}{2})   \sqrt{\nu_n(a,b)/n}]=1-\alpha,
 $$
 where  $\Phi^{-1}$ in the probability statement is the standard normal quantile, so that for $\alpha=0.05$, $\Phi^{-1}(1-\frac{\alpha}{2})\approx 1.96$, and we have
 
 $$\tau_n(a,b) \pm 1.96  \sqrt{\nu_n(a,b)/n}$$
 
 as an approximate $95\%$ confidence interval for the Tversky index $\tau(a,b)$, based on a data set with large $n$.
 \end{cor}
 
  \subsection{Advantages}
 There are two advantages of these explicit formulas.
 
 \begin{adv} They can make the computations on standard error and confidence intervals faster (compared to other methods such as the bootstrap). This advantage may be especially useful in research related to repeated computations of confidence intervals or standard errors, such as with many real or simulated data sets.\end{adv}
   
 \begin{adv} The formulas can be used to plan for the data size to achieve a required standard error $\delta$ for the F  measure estimation, as we discuss below.
   \end{adv}
   
   \begin{cor} Result (ii) of the proposition implies that for any $\delta>0$, if we let
   $$nEZ   \geq \frac{V(\max\{a,b\})}{\delta^2  b }\; or\; n  \geq \frac{V(\max\{a,b\})}{\delta^2  bEZ  }  ,$$
   then the standard error of $\tau_n(a,b)$ satisfies $$\sqrt{\nu(a,b)/n} \leq \delta.$$ 
   \end{cor} 
   
   Since in practice $nEZ$ is estimated by the  data size with $Z=1$, the first formula tells how many observations with $Z=1$ should be recruited.   
   Suppose we know the  parameter   $EZ$  from a preliminary study, then the second formula   may be used to 
  plan for an overall sample size $n$ so that  the standard error of $\tau_n(a,b)$ is at most $\delta$.   
  
This lower bound of the sample size does not involve   $EA$  or the F measure $\tau(a,b)$, and can be used when the new study may use a different prediction rule $A$, making   both $EA$ and $\tau(a,b)$  different from that of the preliminary study. E.g., $A=I(S>0.5)$, where $S$ in the preliminary study is $P_{logistic}(Z=1|X)$ from a logistic regression on some explanatory variables $X$, while as in the new study $S$ may be   $P_{NN}(Z=1|X)$ obtained  from an unspecified neural network, leading to different $EA$ and $\tau(a,b)$.

 \subsection{Example application}
 
 As an application to the previously described orange juice data set, suppose we would like to find a confidence interval for the F measure $\phi_{0.5}=\tau(0.8,0.2)$.  
 
 We first obtain 
 $(n, E_nZA, \tau_n(0.8,0.2),\tau_n(0.8^2,0.2^2))=
 (535,0.535, 0.861, 0.900)$. The formulas in the proposition and Corollary 1 then lead  to the following approximate 95\% confidence interval for the true F measure $\phi_{0.5}$:
 
 $0.861 \pm 1.96\sqrt{ \frac{ 0.900^{-1} -1 +(0.861^{-1}-1)^2}{535(0.535)0.861^{-4} }} =
 0.861\pm 1.96*0.0162=0.861 \pm 0.032.$
 
Suppose we would like to design a new study to have a 95\% confidence interval for $\phi_{0.5}=\tau(0.8,0.2)$ with   a smaller half width $\pm$ 0.02, then how much data will we need?

Note that the required standard error would be  $\delta=0.02/1.96 \approx 0.01$.
Using  the first sample size formula described in Corollary 2 for the case $(a,b)=(0.8,0.2)$ and using $V(\max\{a,b\})=0.2050$ from Proposition 1 Result (iii), we need
 to recruit this number of observations with $Z=1$:
 
 $nEZ \geq \frac{V(max\{0.8,0.2\})}{(0.01)^2   0.2  }=\frac{0.2050}{(0.01 )^2  (0.2)}=10250$.
 
 If we assume that the estimated value   0.615 for $ EZ $ from the previous study is also good for the population in the new study, then  we need an overall sample size
 
 $n\geq 10250/EZ =\frac{10250 }{ 0.615 }=16667$.
 
 So planning for a data size $n=16667$  will guarantee the resulting 95\% confidence interval for the unknown F  measure $\phi_{0.5}$ to have   half width narrower than $\pm 0.02$.

 The sample size planned here is much larger than before, because it is conservative and accommodates maximum standard error caused by any parameters $EA$ and  $\phi_{0.5}$. Consequently, we do not need to assume that the new F measure $\phi_{0.5}$ or the new percentage $EA$ (of purchases predicted to be $Z=1$) need to be similar to the current study. We allow  the use of any unspecified   prediction rule $A$ in the new study, e.g.,   from an unspecified neural network (instead of from logistic regression). The more conservative sample size is an exchange for the flexibility of the prediction method to be used.

 \subsection{Numerical evidences for the proposition}
To verify the asymptotic normality, we consider this model:
$Z\sim Bernoulli(0.5)$, $A=I(S>1)$ where $S|Z\sim N(2.5Z,1)$. We found that the true F measure $\phi_{0.5}=\tau(0.8,0.2)=0.8691099$ based on Monte Carlo method from a million realizations of $(Z,A)$. 
Now we simulate   an iid sample of  $(Z,A)$ with data set size $n=1000$.   We get a sample F  measure $\tau_n(0.8,0.2)$, standard error $se_n\equiv \  \sqrt{\nu_n(0.8,0.2)/n} $, as well as the 95\% confidence interval $\tau_n(0.8,0.2) \pm 1.96 se_n$. Now repeat this for $n_r=10000$ times. We get 
10000 of these quantities 
$(\phi_n, se_n, \phi_n \pm 1.96 se_n)$. We found that the true standard deviation of these
10000 $\phi_n$'s is $sd(\phi_n)=0.01283146$. 

The proposition implies that we can use the standard error $se_n$ to estimate this $sd(\phi_n)$.
The average of 10000 $se_n$'s is actually 0.01280255,
which is indeed very close to the true standard deviation $sd(\phi_n)=0.01283146$.

The proposition implies that the 10000  confidence intervals $\phi_n \pm 1.96 se_n$ will cover the true F0.5 measure $\phi_{0.5}$ about  95\% times. Actually 9455 out of these 10000 intervals indeed cover the true F  measure $\phi_{0.5}$, which gives a percentage 94.55\% being very close to the percentage 95\% suggested by the proposition. 

A histogram of the 10000 estimated   values f0.5=$\tau_n(0.8,0.2)$ for $\phi_{0.5}$  is shown in Figure 1, which shows that these sample values indeed follow a normal-looking distribution. Also, the mean of these 10000 estimated F  values equals 0.8693013, which is very close to the true F  value $\phi_{0.5}=\tau(0.8,0.2)=0.8691099$. 

All these are evidences that support the validity of the proposition.

\includegraphics[width=80mm,height=80mm]{ 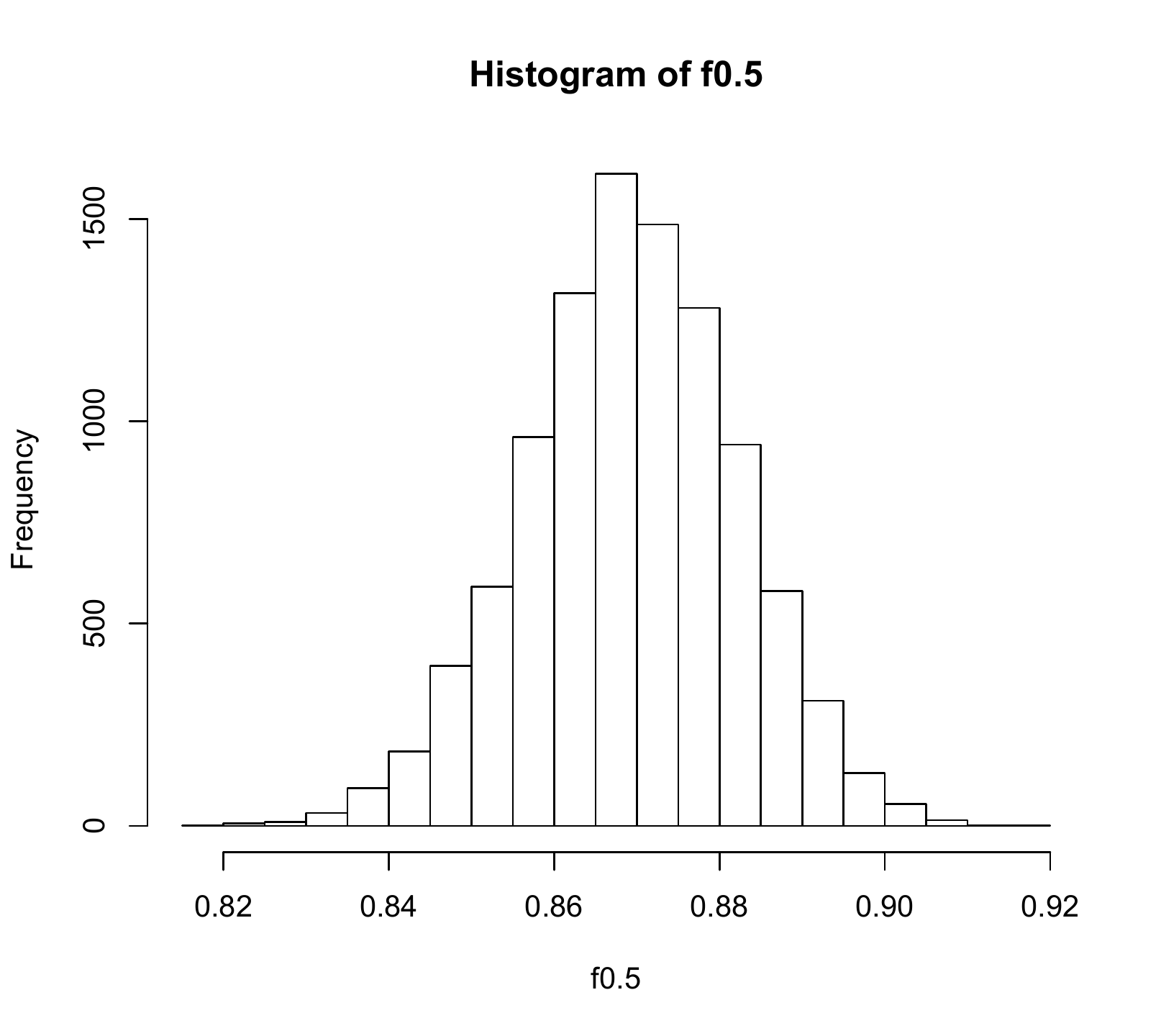}\\

 \section{Proof of the proposition}
   
 For (i): The $\tau_n(a,b)$ can be written as $g(\theta_n)$, which is a smooth function of $\theta_n\equiv E_n[ZA, aA(1-Z)+bZ(1-A) ]^T$. The $\theta_n$, being  the sample average of 2-dimensional iid random vectors, is asymptotically normal due to the central limit theorem: $\sqrt{n}(\theta_n-\theta)$ converges in distribution to $N(0, var[ZA,  aA(1-Z)+bZ(1-A) ]^T)$.
 Then by Taylor expansion, 
 $\sqrt{n}(\phi_n(b)-\phi(b))=\sqrt{n} (g(\theta_n) -g(\theta))\approx g'(\theta) \sqrt{n}(\theta_n-\theta)
$  converges in distribution to $N(0, g'(\theta)var[ZA,  aA(1-Z)+bZ(1-A) ]^T[g'(\theta)]^T)$. Evaluating the derivative and the variance matrix with some algebra leads to Result (i). 

Result (iii) is obtained from (ii) numerically.

For (ii): Note that for $a,b>0$,
$$
\frac{[\tau(a^2,b^2)]^{-1}-1} {[\tau(a ,b )]^{-1}-1} = \frac{a^2EA(1-Z)+b^2EZ(1-A)}{aEA(1-Z)+b EZ(1-A)} \leq \max\{a,b\}.
$$
So 
$$
\nu(a,b)\leq  \frac{[\tau(a ,b )]^4 \{\max\{a,b\}[\tau(a ,b )]^{-1}-1]+ [\tau(a ,b )]^{-1}-1]^2\} }{ EAZ}
$$
$$
=\left(\frac{\max\{aEA,bEZ\}}{EAZ}\right) \frac{[\tau(a ,b )]^4 \{\max\{a,b\}[\tau(a ,b )]^{-1}-1]+ [\tau(a ,b )]^{-1}-1]^2\} }{ \max\{aEA,bEZ\}}.
$$
 
Next we bound the factor
$\max\{aEA,bEZ\}/EAZ$
 in terms of $\tau(a,b)$, using  two relations
$$
\tau(a,b)=\frac{EZA}{(1-a-b)EZA +aEA+bEZ}
$$
and
$$
aEA+bEZ=\max\{aEA,bEZ\}+\min\{aEA,bEZ\}
\geq \max\{aEA,bEZ\}+\min\{a ,b \} EZA.
$$
We obtain
$$\max\{aEA,bEZ\}(EAZ)^{-1}
\leq 
[\tau(a,b)]^{-1}-1+\max\{a,b\}.$$
Therefore
$$
\nu(a,b)\leq    \frac{[\tau(a ,b )]^4 \{\max\{a,b\}[\tau(a ,b )]^{-1}-1]+ [\tau(a ,b )]^{-1}-1]^2\} )[\tau(a,b)]^{-1}-1+\max\{a,b\})}{ \max\{aEA,bEZ\}}.
$$
Simple algebra leads to
$$
\nu(a,b)\leq    \frac{ \tau(a ,b )(1-\tau(a ,b ))(1-(1-\max\{a,b\})\tau(a,b))^2  }{ \max\{aEA,bEZ\}}
\leq  \frac{ \max_{\tau\in (0,1)}\tau(1-\tau )(1-\tau /c)^2  }{  bEZ },
$$
where $c=(1-\max\{a,b\})^{-1}$.
By taking the derivative, one can locate the maximizer to be $\tau=\tau_o$. This leads to the proof.

Q.E.D.

\section*{References}

 \begin{description}
 
 \item Dice, L. R. (1945). Measures of the Amount of Ecological Association Between Species, 
Ecology  26, 297-302.

 \item Elston, R. C., Schroeder, S. R., and Rohjan, J. (1982). Measures of Observer Agreement When Binomial Data Are Collected in Free Operant Situations. Journal of Behavioral Assessment  4, 299-310.
 
 \item Goutte, C. and Eric Gaussier, E. (2005).
A Probabilistic Interpretation of Precision, Recall and F-score, with Implication for Evaluation, 
in D.E. Losada and J.M. Fernandez-Luna (eds) Proceedings of the European Colloquium on IR Resarch (ECIR’05), LLNCS 3408 (Springer),  345–359.

\item Janson, S., and Vegelius, J. (1981). Measures of Ecological Association,  Oecologia, 49, 371-376.

 \item Jiang, W. and Zhao, Y. (2015), On Asymptotic Distributions and Confidence Intervals for LIFT Measures in Data Mining,    Journal of the American Statistical Association   110, 1717-1725.

\item Itzikovitch, R. (2019),
Are We Confident Our Model’s Recall is Precise?\\
https://towardsdatascience.com/are-we-confident-our-models-recall-is-precise-133112a6c407

\item Stine, R. A., Foster, D. P., and Waterman, R. P. (1998), Business Analysis Using Regression: A Casebook, New York: Springer.  

\item Tversky, Amos (1977).  Features of Similarity,Psychological Review  84, 327–352.

\item  Wang, Y., Li, J. , Li, Y.,  Wang, R. and Yang,  X.   (2015),  Confidence interval for  $F_1$ measure of algorithm performance based on blocked 3 × 2 cross-validation , IEEE Trans. Knowl. Data Eng., 27, 651-659.

\item Wikipedia, F score.\\
https://en.wikipedia.org/wiki/F-score

 \end{description}

\end{document}